\documentclass[letterpaper, 10 pt, conference]{ieeeconf}  % Comment this line out if you need a4paper

\IEEEoverridecommandlockouts                              % This command is only needed if 
\usepackage{amsmath,amssymb,graphicx, subfigure,epstopdf,cite,enumerate,booktabs,algorithm,algorithmic,setspace,float,pgfplots}
\usepackage{todonotes, verbatim}

\usepackage[hypcap]{caption}

\usepackage{hyperref}
\hypersetup{
colorlinks=true,
linkcolor=blue,
 filecolor=blue, 
urlcolor=blue,
citecolor=green,
}

\title{\LARGE \bf
Runtime Safety Assurance for Learning-enabled Control of Autonomous Driving Vehicles
}
\author{Shengduo Chen$^{1}$, Yaowei Sun$^{1}$, Dachuan Li$^{1,2,*}$, Qiang Wang$^{2}$, Qi Hao$^{1,2,*}$ and Joseph Sifakis$^{2}$ %and Joseph Sifakis$^{2}$% <-this % stops a space
\thanks{This work is supported by the Shenzhen Fundamental Research Program (No: JCYJ20200109141622964)}% <-this % stops a space
\thanks{$^{1}$S. Chen, Y. Wei, D. Li, Q. Hao are with Department of Computer Science and Engineering, Southern  
        University of Science and Technology, 518055 Shenzhen, China
        {\tt\small \{11860006, sunyw, lidc3 \}@mail.sustech.edu.cn, haoq@sustech.edu.cn}}%
\thanks{$^{2}$D. Li, Q. Wang Q. Hao and J. Sifakis are with Research Institute for Trustworthy Autonomous Systems, 518055 Shenzhen, China
        {\tt\small wangq8@sustech.edu.cn, joseph.sifakis@imag.fr}}%
\thanks{*Corresponding authors: \textit{Dachuan Li, Qi Hao}. S. Chen, Y. Sun and Q. Wang contributed equally to this work}%
}

\begin{document}

\maketitle
\thispagestyle{empty}
\pagestyle{empty}

%%%%%%%%%%%%%%%%%%%%%%%%%%%%%%%%%%%%%%%%%%%%%%%%%%%%%%%%%%%%%%%%%%%%%%%%%%%%%%%%
\begin{abstract}

Providing safety guarantees for Autonomous Vehicle (AV) systems with machine-learning based controllers remains a challenging issue. In this work, we propose Simplex-Drive, a framework that can achieve runtime safety assurance for machine-learning enabled controllers of AVs. The proposed Simplex-Drive consists of an unverified Deep Reinforcement Learning (DRL)-based advanced controller (AC) that achieves desirable performance in complex scenarios, a Velocity-Obstacle (VO) based baseline safe controller (BC) with provably safety guarantees, and a verified mode management unit that monitors the operation status and switches the control authority between AC and BC based on safety-related conditions. We provide a formal correctness proof of Simplex-Drive and conduct a lane-changing case study in dense traffic scenarios. The simulation experiment results demonstrate that Simplex-Drive can always ensure the operation safety without sacrificing control performance, even if the DRL policy may lead to deviations from the safe status. %even in the presence of unverified learning-enabled units.
%The framework is implemented through a case study of lane-changing in dense traffic scenarios,  and simulation experiment results demonstrate that the system can ensure operation safety even if the DRL policy may lead to deviation of safe status, while still achieving desirable control performance. 

\end{abstract}

%%%%%%%%%%%%%%%%%%%%%%%%%%%%%%%%%%%%%%%%%%%%%%%%%%%%%%%%%%%%%%%%%%%%%%%%%%%%%%%%
\section{INTRODUCTION}
As higher levels of autonomy and intelligence of Autonomous Vehicles (AV) are demanded, there is a growing trend of utilizing data-driven machine-learning (ML) techniques (e,g, Deep Reinforcement Learning, DRL) in AV systems, due to their model-free flexibility and experience learning capabilities. The machine-learning based approaches can achieve superior performance in the control of systems with complex dynamics in dynamic environments.

However, the integration of ML components in the system control framework poses a significant challenge to the safety verification of AV systems. The black-box nature of ML units and lack of explicit logic explainability, prevent using conventional formal verification methods at the design time. The control policies of ML units are derived from the training dataset, hence their safety assurance are implicitly determined by the training data. However, there are no applicable metrics and approaches for the validation and verification of datasets, and hence data-driven component. As a result, the lack of safety assurance prohibits the massive application of ML techniques in safety-critical AV systems.

To address the above challenge, in this paper we seek to leverage the \textit{runtime assurance} approach and propose a \textit{Simplex-Drive} framework for the control of AVs. The framework is built upon the basic Simplex architecture \cite{sha2001using,mehmood2021safe}, where the system keeps track of its status at run time, and if necessary, switches the control authority from an unverified performance controller to a proved safe controller to guarantee the runtime safety. When the safe controller is in control, we further incorporate a verified mechanism to enable Simplex-Drive to switch the authority back to the performance controller if safe states can be recovered, so as to restore control performance. In this manner, the system can incorporate ML-based components without requiring correct-by-construction properties at the design stage, while still ensuring provable safety of the overall composite system at run time. Taking the advantage of the Simplex-Drive architecture, this study focuses on the runtime assurance design of Reinforcement Learning-enabled AV controllers with applications to lane-changing scenarios. The primary contributions of this paper are as follows:

\begin{itemize}
    \item Development of the novel Simplex-Drive architecture for the control of AVs, which can provide provable runtime safety assurance for control systems containing data-driven components.
    \item Development of a control module with runtime assurance for lane-changing control of AVs, which consists of an unverified DRL-based performance controller and a verified Optimal Reciprocal Collision Avoidance (ORCA, \cite{orca}) based safe controller. We provide formal safety specifications for lane-changing control tasks and a set of formal proofs of the correctness of the overall framework.
    \item Demonstration of the advantages of Simplex-Drive through comparative evaluations based on a number of simulated lane-changing scenarios. The open-source code of Simplex-Drive is available at:\\ 
    \href{https://github.com/625160928/Safety\_RL\_VO}{https://github.com/625160928/Safety\_RL\_VO}.
\end{itemize}

The remainder of the paper is organized as follows: Section II reviews the related work. Details of the proposed Simplex-Drive framework are presented in Section III, followed by experimental results and analyses in Section IV. The paper is concluded in Section V.

\section{RELATED WORK}

%\subsection{Selecting a Template (Heading 2)}

%First, confirm that you have the correct template for your paper size. This template has been tailored for output on the US-letter paper size. 
%It may be used for A4 paper size if the paper size setting is suitably modified.

The integration of deep reinforcement learning (DRL) into AV applications have been drawing a lot of attention in recent years. Taking advantages of DRL, AVs can efficiently derive control policies and achieve better performance than traditional methods that rely on explicit system models and pre-designed logic \cite{DQN_Drive} \cite{EndtoEndDrive} \cite{Learningtodrive} \cite{AdaptiveBG} \cite{HRLDrive} \cite{AttentionHDRL}. Specifically, a modified Q function approximator \cite{con_DQN} is utilized to generate continuous lateral acceleration actions for AVs from the continuous state space \cite{DQN_Drive}. However, the proposed method has been applied to a simple situation assuming that other vehicles only have the longitudinal actions and never change their lanes. For the longitudinal control, a pre-defined Intelligent Driver Model (IDM) has been developed to generate acceleration actions. The later work \cite{DrivingDA} also uses a neural network to approximate the Q function which can generate continuous actions for longitudinal control. When a lane change decision is make, the network generates actions to adjust longitudinal distances and then a fifth degree polynomial curve is executed to achieve a complete lane change. To control the AV along both lateral and longitudinal directions at the same time, an Asynchronous Learning strategy (A3C) \cite{a3c} is developed with a network architecture of three convolutional layers and one fully connected layer to generate control actions for AV in race driving \cite{EndtoEndDrive}. However, such a network can only handle the discrete action space which may result in poor smoothness in trajectories. Other DRL algorithms have also been applied to control AV for lane-change and lane-merge scenarios \cite{Learningtodrive}, \cite{AdaptiveBG} and \cite{2020EndtoEndMR}. However, none of these approaches can provide theoretical proofs of the safety  assurance of generated actions.

The idea of leveraging runtime assurance to address the safety issue of complex and unverified controllers was initially proposed in \cite{sha2001using}. The proposed Simplex architecture has been widely adopted in many safety-critical applications such as flight control systems \cite{schierman2015runtime}, drone surveillance systems \cite{desai2019soter}, collision avoidance of mobile robots \cite{phan2017collision} and automatic aircraft taxiing control systems \cite{cofer2020run}. Such domain applications lead to extensions of Simplex focusing on the design of the safety verification/monitoring mechanism and switching logic. The original Simplex leverages Lyapunov based approaches, which can only be applied to continuous systems. Therefore, \cite{bak2010hybrid, bak2014real} utilize a set of hybrid automaton and reachability analysis techniques to design the Simplex switching logic for hybrid systems. However, previous Simplex-based frameworks do not focus on any specific controllers and treat all unverified components as black boxes. Moreover, the lack of a reverse switch mechanism prevents the system from restoring control performance. Recently, the Neural Simplex architecture \cite{phan2020neural} has been proposed to incorporate RL-based components to the Simplex framework, along with a bi-directional switch logic. Inspired by Neural Simplex, we provide runtime assurance for an AV control framework with a DRL-based performance controller in this study.

\section{SIMPLEX-DRIVE FRAMEWORK WITH RUNTIME ASSURANCE}

\subsection{Overview of Simplex-Drive framework} 
\begin{figure}[htp]
    \centering
    \vspace{0.2cm}
    \includegraphics[width=0.485\textwidth,clip]{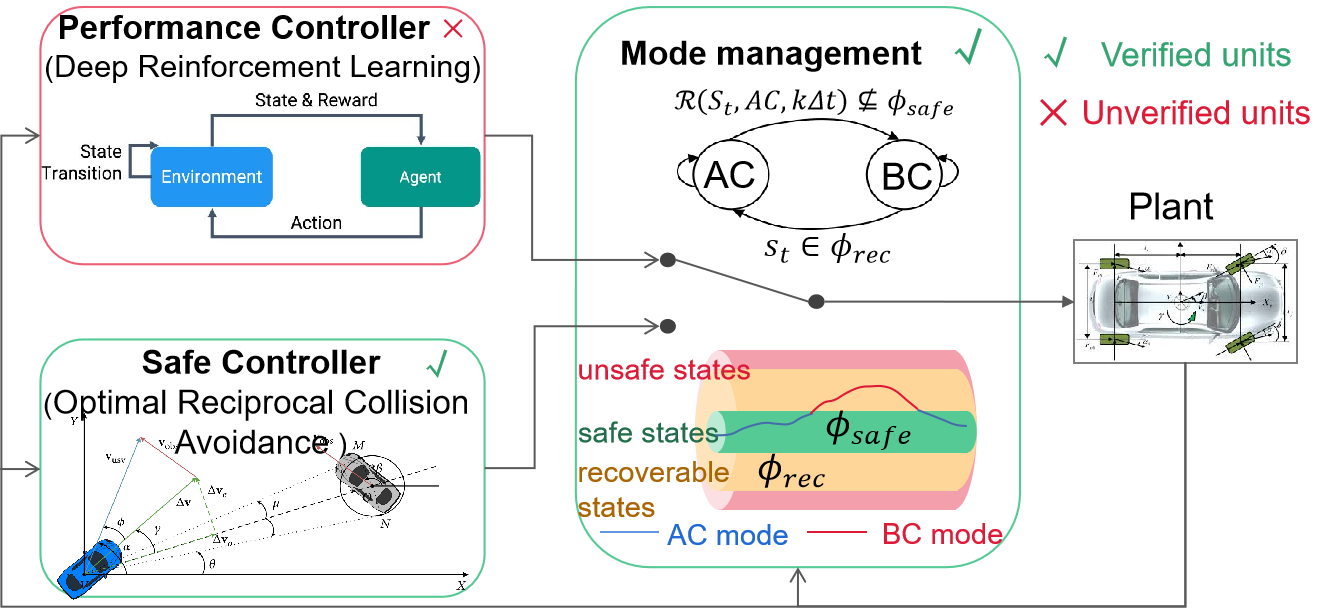}
    \caption{The system diagram of the proposed Simplex-Drive framework for the machine learning based autonomous vehicle control with safety assurance. AC is used to achieve desired performance and BC for safety assurance. MM monitors the operation condition and switches between these two control authorities.}
    \label{fig:framework}
\end{figure}

The overall Simplex-Drive framework is illustrated in Fig. \ref{fig:framework}. The framework consists of three major components: namely the \textbf{Advanced performance controller (AC)}, the  \textbf{Baseline safe controller (BC)} and the \textbf{Mode management unit(MM)}.

The dynamics of the controlled vehicle is given by $s_{t+1}=f(s_t,u_t)$, where $s_t\in\mathcal{S}$ and $u_t\in\mathcal{U}$ denote the sate and control at time step $t$, respectively. We denote with $CM_{t}\in\{\verb|AC,BC|\}$ as the control mode determined by MM unit at time step $t$, where $\verb|AC,BC|$ indicate the AC (BC) controller has the control authority, respectively. Let $u^{AC}_{t}, u^{BC}_{t}$ be the control control output of AC and BC controller.

The AC is data-driven and maps the current sate into a control action: $u^{AC}_{t+1}=\pi_{\theta}^{AC}(s_t)$. In this study, AC can be obtained by training the DRL policy. In contrast, the BC ($u^{BC}_{t+1}=\pi^{BC}(s_t) $) is required to be designed with provable safety guarantee. In the case study of this paper, we design a BC based on the ORCA principle\cite{orca}, which is proven to preserve collision avoidance property. The MM monitors the state $s_t$ at runtime and switches control authority from AC to BC if $u^{AC}_t$ will lead the system to an \textit{unrecoverable state}  (i.e. a state where the system cannot guarantee safety property) at some time step in the future. When $CM_t=\verb|BC|$, the MM can also switches the mode back to $\verb|AC|$ to resume AC control if the system returns to a \textit{recoverable state}. Both MM and BC are proven to guarantee safety properties, while AC is not required to be verified correct-by-construction.

\subsection{The Deep Reinforcement Learning Controller}

For lane-change scenarios, the process of generating vehicle control commands can be formulated as a Markov Decision Process (MDP) $<S,A, P, r, S, \gamma>$. We use the terms “state” to describe the status of the surrounding environment, which can be observed by the agent. Without loss of generality, we assume a continuous action space $a \in A$ for the agent. Let $\pi_{\theta}(a|s)$ be the policy distribution with learnable parameters $\theta$, and $P(s_{t+1}|s_t, a_t)$ the transition probability that measures how likely the environment transitions to $s_{t+1}$ given an action by $a_t \sim \pi_{\theta}(·|s_t)$. After the transition to $s_{t+1}$, the agent receives a discounted reward $r(s_t, a_t, s_{t+1})$. The ${\gamma}^i r_{t+i}$ is the discounted return. The objective of the DRL is to solve the above MDP by learning a policy that maximizes the expected total discounted rewards.

\begin{figure}[htp]
\vspace{0.1cm}
    \centering
    \includegraphics[width=0.45\textwidth,clip]{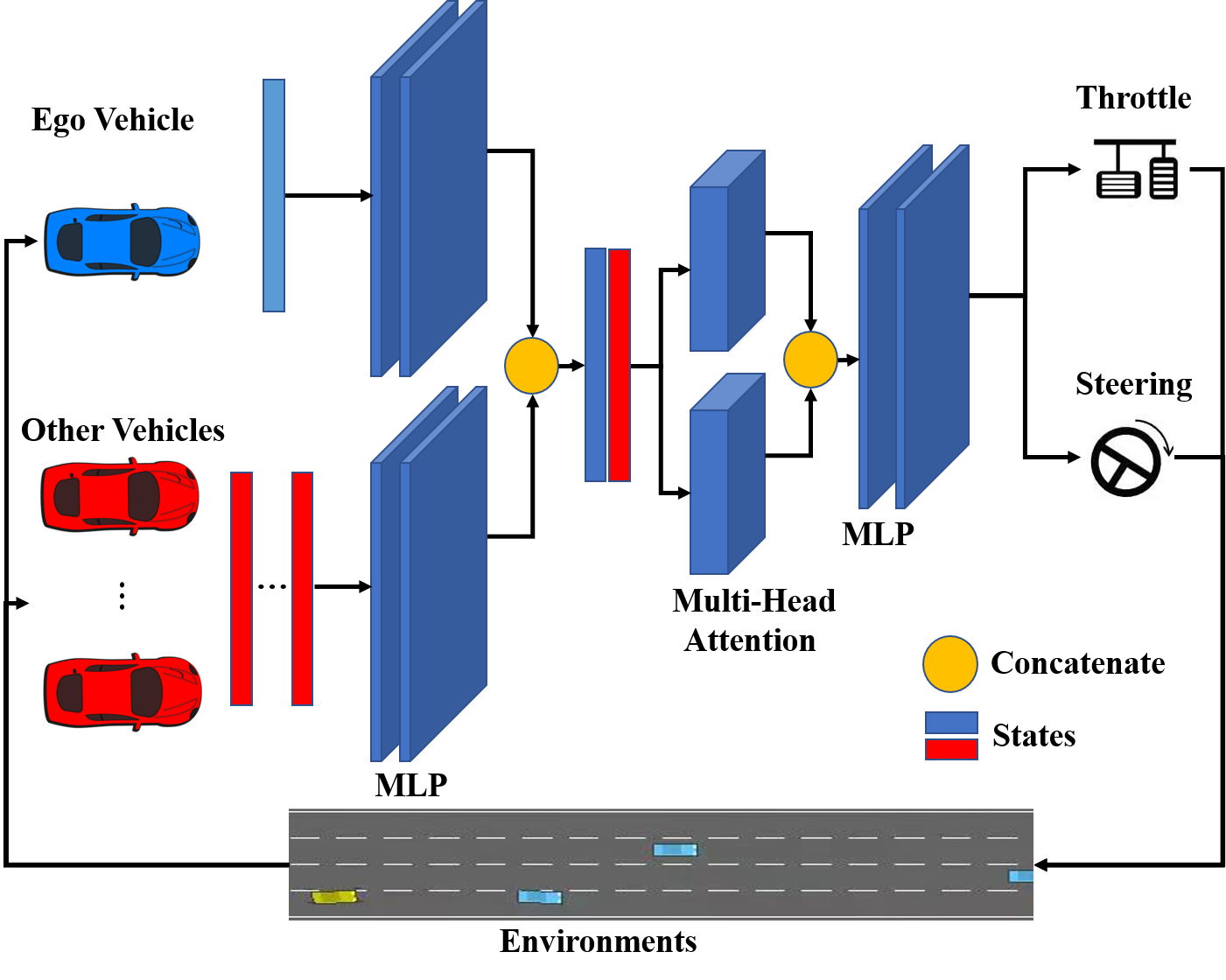}
    \caption{The network architecture of the proposed deep reinforcement learning based controller. The first two MLP layers are used for environment representation; the multi-head attention layer is used for focusing on nearest vehicles; the last two MLP layers are used for policy formulation.}
    \vspace{-0.8cm}
    \label{fig:network}
\end{figure}

\subsubsection{The Network Architecture}

To design the lane-changing control approach in autonomous driving applications, we develop a model-free policy with a multi-head attention network \cite{attention} to map the state to continuous actions. The input states consisting of the position, velocity and heading information of the ego vehicle are fed into two fully-connected layers multi-layer perceptron (MLP) network with 128 neurons and Relu activation functions. The states of surrounding vehicles (which consists of same elements as the ego-vehicle) are sent to another MLP network with the same architecture. Then the concatenated features are sent to a multi-head attention module followed by another MLP with 2 fully connected layers that have 256 neurons. We build the multi-head attention network based on the framework proposed by \cite{attention}. The output layer has two units with tanh activation functions that output the throttle and yaw control commands as actions. The overall network architecture is shown in Fig. \ref{fig:network}. The policy is trained under the Actor-Critics scheme.

\subsubsection{Design of DRL-based Controller}

The reward settings for training the DRL algorithm are set as follows. Reaching the target lane and pose keeping are encouraged by positive rewards, while hazardous behaviors such as collisions and violations of lane borders are penalized by negative rewards. To take the advantage of the DRL-based controller to control the AV more efficiently, we incorporate an additional efficiency-related reward term r1 and r3 to encourage the agent to complete lane-changing in a time-efficient manner. The ego vehicle receives r1 when reaching the target lane and r2 when collision occurs. r4, r5 are the rewards used to constrain the the vehicle heading.

\iffalse
\begin{figure}[htp]
    \centering
    \includegraphics[width=0.48\textwidth,clip]{figure/highway.PNG}
    \caption{The highway environment for training.}
    \label{fig:environment}
\end{figure}
\begin{figure}[htp]
    \centering
    \includegraphics[width=0.48\textwidth,clip]{figure/carla-manual-control.jpg}
    \caption{The carla environment for testing.}
    \label{fig:carla}
\end{figure}
\fi

\subsection{The Optimal Reciprocal Collision Avoidance-based Safe Controller}

\begin{figure}[htp]
    \centering
    \resizebox{\linewidth}{!}{
   $\begin{array}{cc}
   % Requires
    \subfigure[]{\includegraphics[height=.3\textwidth]{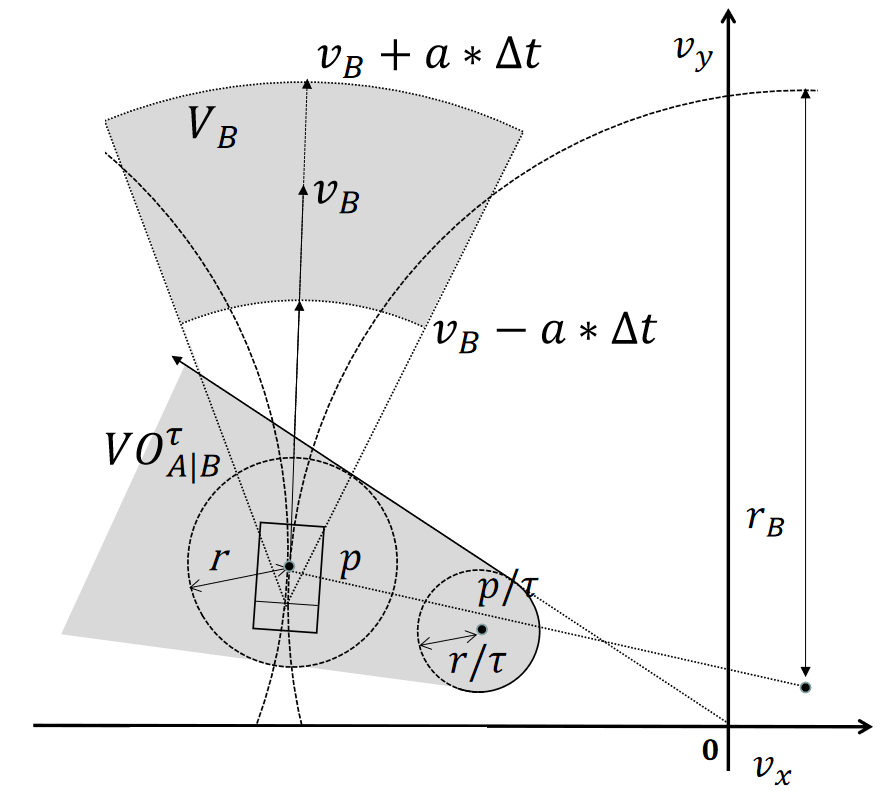}} &
   \subfigure[]{\includegraphics[height=.3\textwidth]{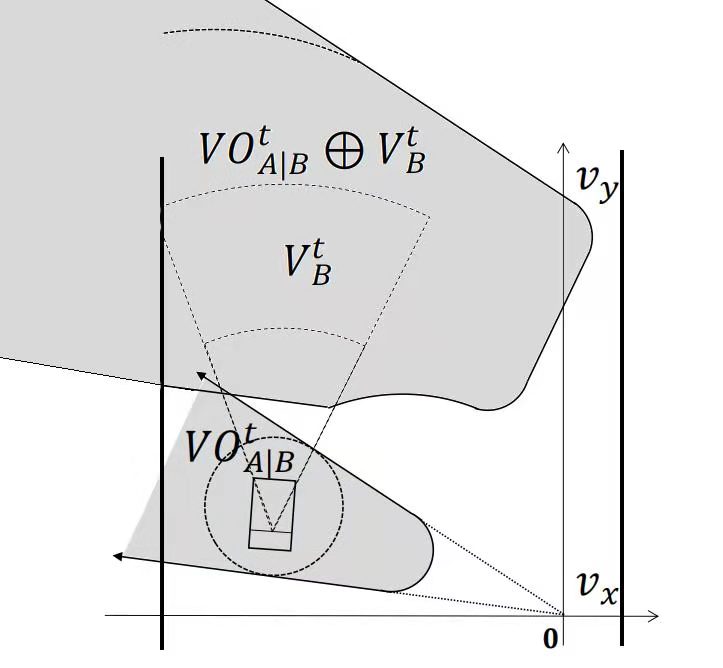}} \\
 
   \end{array}$
   }
\end{figure}
\begin{figure}[htp]
    \centering
    \resizebox{\linewidth}{!}{
   $\begin{array}{cc}
   % Requires
    \subfigure[]{\includegraphics[height=.3\textwidth]{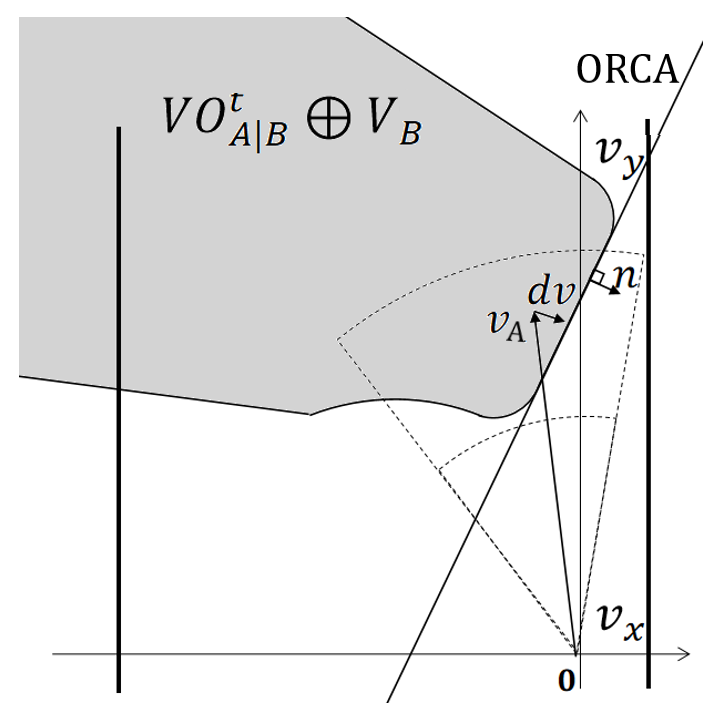}} &
   \subfigure[]{\includegraphics[height=.3\textwidth]{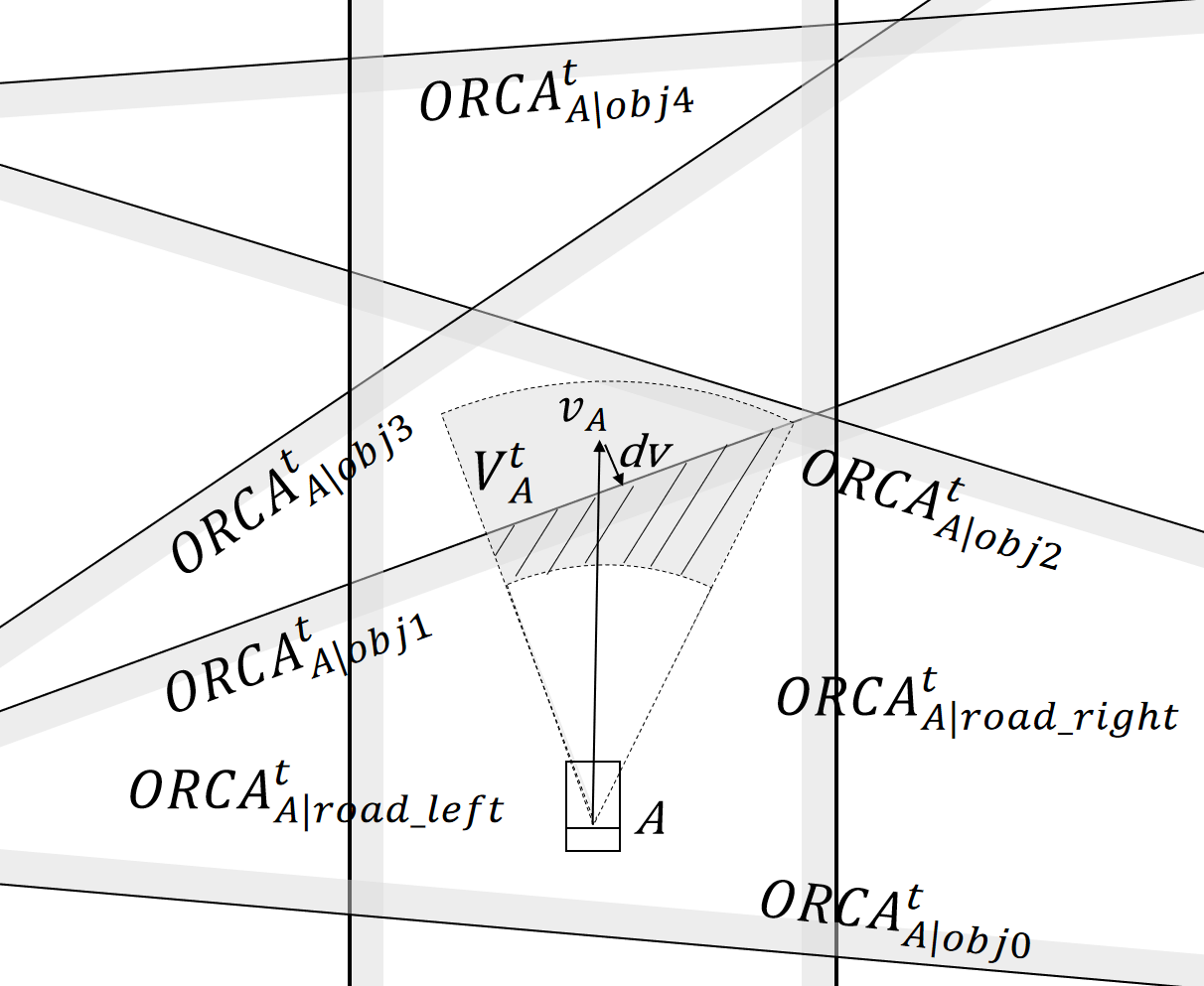}} \\
   \end{array}$
   }
   \caption{The Proposed ORCA-Drive safe controller. (a) The reachable velocity set $V_B$ and velocity obstacle set $VO^{\tau}_{A|B}$ for the ego vehicle A at the origin and the obstacle vehicle B denoted by a rectangle. (b) The resultant velocity obstacle area $VO^{\tau}_{A|B}$ for the ego vehicle A at the origin within a period of $\Delta t$. The computation of the optimal velocity increment $dv$ for the ego vehicle at the origin given (c) a single obstacle vehicle and (d) multiple obstacle vehicles, respectively.}
   
   \label{fig:orca}
   \vspace{-0.2cm}
\end{figure}

The Optimal Reciprocal Collision Avoidance (ORCA) \cite{orca} approach was originally proposed for multi-robot collision avoidance and it has been theoretically proved to be able to preserve the safety property. Therefore, we utilize the ORCA as the baseline safe controller in our system to guarantee safety. Since the ORCA is originally designed for omni-directional robot, we extend the ORCA to the control of Ackermann dynamic vehicles.
As shown in Fig. \ref{fig:orca} (a), surrounding vehicles related to the ego-vehicle movements are regarded as obstacles. Considering the ego vehicle as $A$ with velocity $v_A$ and obstacle as $B$ with $v_B$, we firstly calculate the velocity reachable set of the obstacle $V_B$, where the upper bound and lower bounds are calculated with the maximum acceleration $a$ in a predicted time interval $\Delta t$. The left and right bounds are calculated with the maximum steering angle and the $r_B$ is the minimum steering radius of the obstacle vehicle with the Ackermann dynamic model.
$VO^{\tau}_{A|B}$ is the velocity obstacle for ego vehicle induced by $B$ for a time window $\tau$ with the current relative velocity w.r.t the obstacle vehicle (in a typical AV system configuration, velocities of surrounding vehicles can be measured using onboard sensors such as radars). $P$ is the relative position between the ego and obstacle vehicles; $\tau$ is set to 2 in our configuration; $r$ is the radius of the minimum bounding circle of the vehicle.

With the velocity obstacle area $VO^{\tau}_{A|B}$ at the current relative obstacle vehicle speed, it is able to generated the velocity obstacle area for a period of $\Delta t$ in the future by calculating the Minkowski Sum $VO^{\tau}_{A|B} \bigoplus V_B$, as shown in Fig. \ref{fig:orca} (b).
According to the original ORCA algorithm, the optimal velocity to avoid collision is generated by adding minimal velocity change $d_v$ to the current ego vehicle velocity $v_A$. In order to address the Ackermann vehicle model constraints, the actions of the ego vehicle should also satisfy the vehicle kinematic constraints. Therefore, we also calculate the velocity reachable set $V_A$ of the ego vehicle. Then the generated velocity $v_{opt} = v_A + d_v$ is checked to make sure it is within $V_A$. The action generation processes of single and multiple obstacle vehicles are illustrated in Fig. \ref{fig:orca} (c) and the Fig. \ref{fig:orca} (d) respectively. Details of the extended ORCA (namely ORCA-Drive) algorithm are shown in Algorithm \ref{alg:A}.

\vspace{-0.2cm}
\begin{algorithm}[h]
    \caption{ORCA-Drive Safe Controller}
    \begin{algorithmic}[1]
    \FOR{ Every prediction period t}
        \STATE Initialize the ORCA plane list $L$
        \FOR{each vehicle $i \in$ other vehicles}
            \STATE Calculate the velocity reachable set $V^t_i$ for vehicle $i$.
            \STATE Calculate VO area $VO^t_{A|i}$ for ego-vehicle $A$.
            \STATE Calculate Minkowski Sum $CA^t_{A|i}(V^t_i) = V^t_i \bigoplus VO^t_{A|i}$.
            \STATE Calculate minimal velocity change $dV_i$ that adjust current velocity $V_A$ to leave $CA^t_{A|i}(V^t_i)$.
            \STATE Generate the ORCA plane $P_i = [L_i, dV_i]$ and store $P_i$ in $L$.
        \ENDFOR 
        \STATE Generate velocity reachable set $V_A$ of the ego vehicle under kinematic constraint.
        \STATE Generate Safe velocity set $U_s$ of the ego vehicle with ORCA plane list $L$.
        \STATE Calculate the optimal velocity set $U_{opt} = U_A \cap U_s$ and get optimal velocity change $dV_{opt}$.
        \IF{$U_{opt} = \varnothing $} 
            \FOR{each sampled $U_n \in U_A$} 
                \STATE Calculate weight $W_n$ += $W_{orca} \times$ distance $d_{orca}$ to ORCA plane for all plane.
                \STATE Calculate weight $W_n$ += $W_{current} \times$ distance $d_{current}$ to current speed.
            \ENDFOR
            \STATE $V_{opt}$ is the $U_n$ with highest weight.
        \ELSE
            \STATE Get the optimal speed with $V_{opt} = V_{current} + dV_{opt}$.
        \ENDIF 
    \ENDFOR
    \end{algorithmic}
    \label{alg:A}
\end{algorithm}

\subsection{Switching logic}
The safety property during operation can be formulated as a safe set: $\phi_{safet}\subseteq\mathcal{S}$, indicating the set of vehicle's states that satisfies safety properties (i.e. safe spacing, traffic rule obedience). At runtime, the system is required to always stay within $\phi_{safe}$. We define an additional recoverable region $\phi_{rec}$:

\noindent\textbf{Definition. 1 (recoverable region)}: The recoverable region is the set of states $s\in\phi_{rec}$ that satisfy the following properties:

\noindent(1) $\mathcal{R}(s,\verb|AC|,\Delta t)\subseteq \phi_{safe}$; (2) $\mathcal{R}(s,\verb|BC|,\infty)\subseteq\phi_{safe}$; (3) $\mathcal{R}(\mathcal{R}(s,\verb|AC|,\Delta t),\verb|BC|,\infty)\subseteq\phi_{safe}$;

\noindent where $\mathcal{R}(s_0,MC,\Delta t)$ indicates the reachable sets of the vehicle's sates within a time interval $\Delta t$ under control mode $MC$, starting from a initial state $s_0$. Therefore, $\phi_{rec}$ indicates a region from which the BC can always steer the system to safe region $\phi_{safe}$ no matter which controller is in charge.    

The switching logic of the MM unit can be modeled as a finite-state automaton with two control locations (Fig. ): $\verb|AC|$ and $\verb|BC|$, indicating the AC or BC is in control, respectively. The switching condition is given by:
\begin{equation}
    	CM_{t+\Delta t}=\left\{
	\begin{array}{lr}
		\!\verb|BC|\! & \! CM_t=\verb|AC| \land \mathcal{R}(s_t,\verb|AC|,k\Delta t)\nsubseteq \phi_{safe}\\
		\!\verb|AC|\! & \! CM_t=\verb|BC| \land s_t\in\phi_{rec}\\
		\!CM_t\! & \! otherwise
	\end{array}
	\right.
	\label{eq:switch}
\end{equation}
In our application, the MM unit checks the sate of the system and makes decisions every $\Delta t$. We use bounded reachability analysis \cite{frehse2011spaceex} to predict the set of vehicle's reachable states and check for deviation from the safe region (line 1, Eq. \ref{eq:switch}).  
\subsection{Runtime Assurance of Simplex-Drive}
 The correctness of the overall Simplex-Drive can be verified by the following theorem: 

\noindent \textbf{Theorem 1}:Denoting $\mathcal{R}(s_0)$ as the set of reachable states from $s_0$, 
then $\forall s\in\mathcal{R}(s_0)$, the following invariant holds:

$(MC=\verb|BC|\land s\in\phi_{safe}) \lor (MC=\verb|AC|\land \mathcal{R}(s,\verb|AC|,\Delta t)\subseteq\phi_{safe})$

if the following conditions hold:

\noindent(1) $s_0\in\phi_{safe}$;

\noindent(2) $\Delta t > T_{AC} \land \Delta t > T_{BC} $;

\noindent(3) $\mathcal{R}(\phi_{safe},\verb|BC|,\infty)\subseteq\phi_{safe}$;

\noindent(4) $\mathcal{R}(\phi_{rec},\verb|AC|,k\Delta t)\subseteq\phi_{safe}, k\in\mathbb{N}, k>1$

where $T_{AC}$ and $T_{BC}$ denote the control interval of AC and BC.

\textit{Proof}. Suppose the invariant holds at time step $t$, the theorem is proved by induction on every consecutive interval of the MM unit:

\noindent(1) If $MC_t=\verb|BC|$ and there is no mode switch, all future states satisfy the invariant according to condition (3) of Theorem 1;

\noindent(2) If $MC_t=\verb|BC|$ and $\verb|BC|\rightarrow\verb|AC|$ at time $t$, this implies $s_t\in\phi_{rec}$. According to condition (4), $\mathcal{R}(s_t,\verb|AC|,k\Delta t)\subseteq\phi_{safe}, k>1$, and the invariant holds for the next interval;

\noindent(3) If $MC_t=\verb|AC|$ and there is no mode switch, this implies
$\mathcal{R}(s_t,\verb|AC|,k\Delta t)\subseteq\phi_{safe} \Rightarrow \mathcal{R}(s_{t+\Delta t},\verb|AC|,\Delta t)\subseteq\phi_{safe}$ for the next time interval

\noindent(4) If $MC_t=\verb|AC|$ and $\verb|AC|\rightarrow\verb|BC|$ at time $t$, this implies $\mathcal{R}(s_t,\verb|AC|,k\Delta t)\nsubseteq\phi_{safe}$. However, as we have $\mathcal{R}(s_{t-\Delta t},\verb|AC|,2\Delta t)\subseteq\phi_{safe}$  (since $MC_t\neq\verb|AC|$ otherwise) $\Rightarrow \mathcal{R}(s_t,\verb|AC|,\Delta t)\subseteq\phi_{safe}$. According to condition (2), the BC will execute at least once during the next interval. Therefore, we have $\mathcal{R}(s_t,\verb|BC|,\infty)\subseteq\phi_{safe}$ according to condition (3) and thus the invariant holds. $\square$

\begin{table*}[htpb]
\vspace{0.2cm}
\caption{A PERFORMANCE COMPARISON BETWEEN THE PROPOSED SIMPLEX-DRIVE APPROACH AND OTHER METHODS}
\vspace{-0.2cm}
\begin{center}
\centering
\resizebox{0.9\linewidth}{!}{
\begin{tabular}{cccccc}
\hline
\hline
\textbf{Density = 1 (vehicle/lane)}   & Target Lane Rate & Collision Rate & Avg\_Speed (m/s) & Min\_Dis (m) & Avg\_Min\_Dis (m)  \\
\hline
\textbf{ORCA—Drive}          &90\%  & 0  & 17.44  &19.22  &24.80    \\
\textbf{AVO}           & 63\%  &35\%  &18.96  &5.35  &12.73     \\
\textbf{Attention PPO} & 97\%            & 27\%           & 19.44      & 8.19     & 16.31           \\
\textbf{DQN \& IDM}    & 100\%                & 36\%           & 19.87      & 7.55     & 10.87           \\
\textbf{Simplex-Drive(Ours)}          & 95\%             & \textbf{0}        & 19.36       & 19.17   & 21.32    \\
\hline
\textbf{Density = 1.5 (vehicle/lane)}   & Target Lane Rate & Collision Rate & Avg\_Speed (m/s) & Min\_Dis (m) & Avg\_Min\_Dis (m) \\
\hline
\textbf{ORCA—Drive}          & 77\%             & 0              & 14.96      & 10.62    & 20.52            \\
\textbf{AVO}           & 33\%             & 72\%           & 18.31      & 4.74     & 10.70            \\
\textbf{Attention PPO} & 92\%             & 73\%           & 19.44      & 5.32     & 12.35              \\
\textbf{DQN \& IDM}    & 100\%                & 95\%           & 19.73      & 4.55     & 8.87              \\
\textbf{Simplex-Drive(Ours)}          & 89\%             & \textbf{0}              & 17.91      & 10.22     & 19.36            \\
\hline
\textbf{Density = 2 (vehicle/lane)}   & Target Lane Rate & Collision Rate & Avg\_Speed (m/s) & Min\_Dis (m) & Avg\_Min\_Dis (m)  \\
\hline
\textbf{ORCA—Drive}          & 56\%  & 0  & 15.27  & 8.30  & 14.09      \\
\textbf{AVO}           & 30\%  & 94\%  & 17.66  &4.56  &8.70      \\
\textbf{Attention PPO} & 87\%             & 83\%           & 19.44      & 5.32     & 12.35             \\
\textbf{DQN \& IDM}    & 100\%                & 100\%           & 19.76      & 3.95     & 6.81              \\
\textbf{Simplex-Drive(Ours)}          & 85\%             & \textbf{0}              & 17.80      & 7.23     & 15.78           \\
\hline
\hline
\end{tabular}}
\end{center}
\label{result}
\vspace{-0.4cm}
\end{table*}

It is proved in \cite{orca} that the ORCA-based BC controller can guarantee safety properties. Therefore, Theorem. 1 guarantees that if the specified conditions hold, all reachable states of the vehicle system always stay within the safe region, no matter which mode the system is currently in. 

\section{EXPERIMENT RESULTS}

To evaluate the effectiveness and performance of the pro- posed Simplex-Drive control framework, we conducted extensive experiments in the highway-env simulation environment \cite{highway-env} with different levels of complexity and operation conditions.

\subsection{Experiment Setup}

We developed a simulated environment based on highway-env simulator \cite{highway-env} for the training of the DRL-based AC. Various experiments are also conducted using this simulated environment to verify the effectiveness and safety of our framework. In the experiments, we focus on dense traffic scenarios with 3 lanes (the lane width are assumed to be fixed). The controlled ego vehicle (with initial speed $V_{ego}$) and the obstacle vehicles (with initial speed $V_{other}$) are assumed to have the same size.

\subsection{Comparative experiments}

We compared the proposed architecture with standalone ORCA-Drive controller, AVO controller \cite{AVO}, Attention PPO controller as well as a controller with DQN and IDM \cite{2020EndtoEndMR} in scenarios with different degrees of complexity. The AVO controller utilizes acceleration-velocity obstacle that extend the VO with acceleration constraints to calculate the velocity policy. The DQN with an IDM controller operates within a similar multi-level framework but utilizes DQN to generate high-level decisions and utilize the IDM module for vehicle control. We compared the performance of these approaches in terms of safety and efficiency-related metrics.

The configurations of parameters used in the experiments are shown in TABLE \ref{para}.

\begin{table}[htpb]
\vspace{-0.1cm}
\caption{PARAMETER SETTINGS FOR SIMULATIONS}
\vspace{-0.2cm}
\begin{center}
\centering
\resizebox{0.98\linewidth}{!}{
\begin{tabular}{llll}
\hline
Term                  & Value       &Term                  & Value \\
\hline
Vehicle Length  &  5.0m  &Vehicle Width                  & 2.0m\\
$V_{max}$  &  20m/s  &$V_{other}$                 & 15m/s\\
$Max_{step}$  &  $200$  & $V_{ego}$                  & 20m/s\\
$acceleration_{ego}$  &  [-5 $m/s^2$, 5 $m/s^2$]  &$Steering_{ego}$        & [-$\pi$/6,$\pi$/6]\\
$\lambda_1$             & -0.2   & $\lambda_2$           & 0.2 \\
r1 & 2 & r2 & -5   \\
r3 & $v_{current}-v_{target}$   &r4   & $cos(\theta)$\\
$V_{target}$ & 20 &r5  &-2\\
Lane width                  & $2.5m$          & $\Delta t$                  & 0.5s\\
$W_{orca}$          & 0.7      &$W_{current}$          & 0.3\\
\hline
\end{tabular}}
\end{center}
\label{para}
\vspace{-0.8cm}
\end{table}

\begin{figure*}[th]
\vspace{0.1cm}
    \centering
    \resizebox{\linewidth}{!}{
    \includegraphics[width=0.8\textwidth,clip]{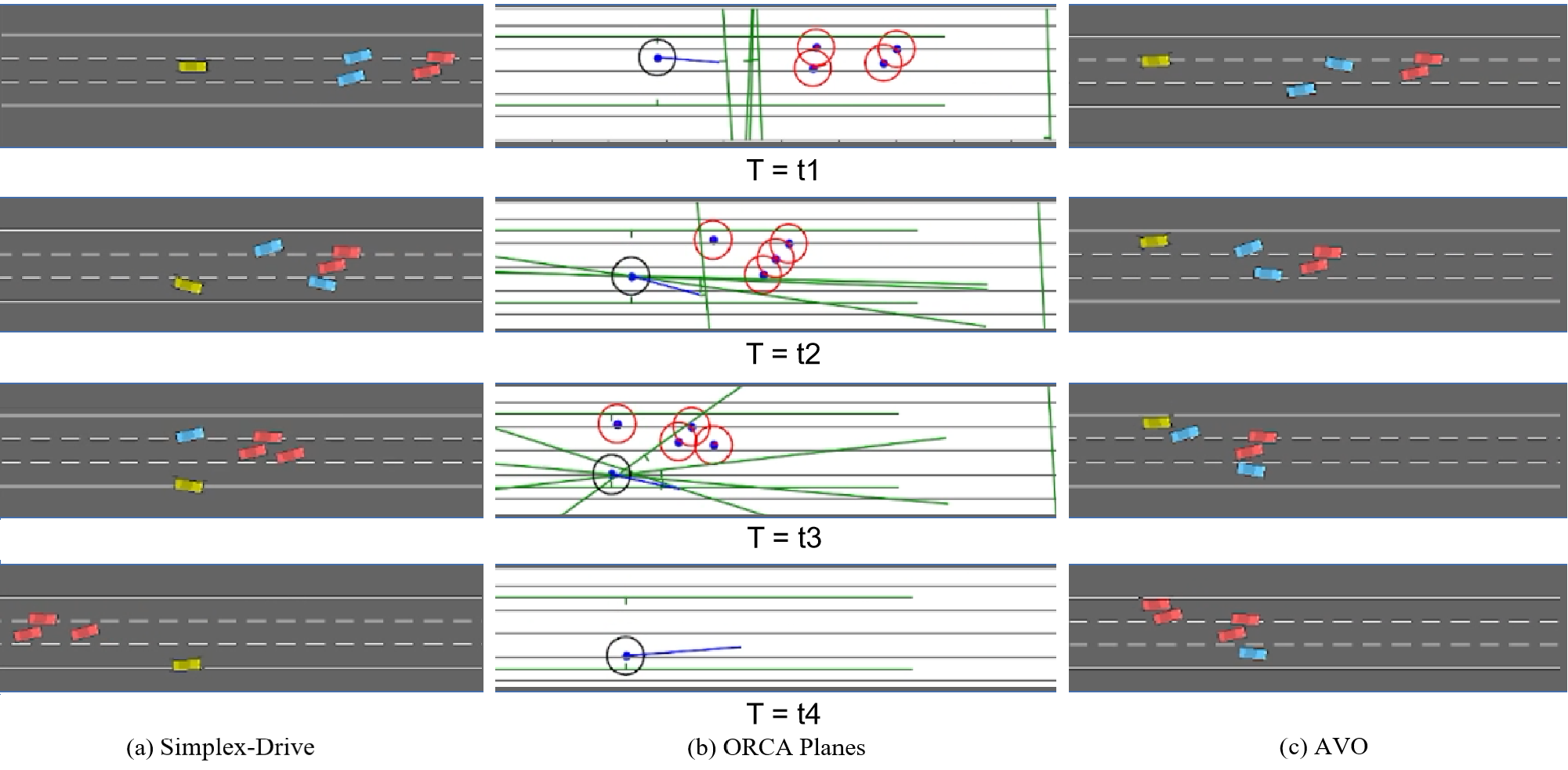}
    \vspace{-0.1cm}
   }
   
   \caption{An illustration of the maneuvers of the ego-vehicle using the Simplex-Drive and ORCA control policies respectively in the presence of traffic accidents. (a) The snap shots of the poses of the ego, crashed, normal vehicles, marked by yellow, red, and blue colors, using the proposed Simplex-Drive controller. (b) ORCA regions of the environment for the Simplex-Drive controller, where the obstacle vehicles and ORCA planes are represented by red circles and green lines respectively.  (c) The snap shots of the poses of ego, crashed, and normal vehicles using the AVO controller for the same scenario as a comparison.}
   \label{orca_sim}
   \vspace{-0.8cm}
\end{figure*}

\subsection{Evaluation Metrics}
%To verify the performance, we utilize several evaluation metrics to indicate the performance of the controller. 
The evaluation metrics used in the experiments include the target lane rate and collision rate, indicating whether the ego vehicle reaches the target lane before the epoch ends and whether the maneuver results in collisions, respectively. The Avg\_Speed is the average  vehicle speed at each time step which indicates the efficiency of lane-changing control. In addition, we utilize the minimal distance (Min\_Dis) to other vehicles and its average during the epoch (Avg\_Min\_Dis) to demonstrate the degree of conservativeness of the controllers. 
We use the vehicle density metric to define the complexity of the environment and the traffic density. The vehicle density is defined as the average number of vehicles on each lane in the near distance. We test the controllers with scenarios with densities of 1, 1.5 and 2. Each density level is tested for 50 times and the results are averaged.
\vspace{-0.2cm}
\subsection{Simulation Results}

The simulation results are shown in TABLE \ref{result}. It can be seen that the ORCA-Drive safe controller can always guarantee the safety with 0 collision during all the experiment trials. Compared with DQN and IDM-based controllers that generate continuous control, the Attention PPO controller can deal with the congested environment with better performance and maintain larger minimal spacing to surrounding vehicles. As expected, although the controller with DQN and IDM can achieve smoother trajectories, the lag between the high-level decision module and low-level control module result in delays when reacting to contingencies. The ORCA-Drive safe controller normally keeps a relative larger distance from other vehicles, compared with other controllers, and such a conservative policy generated by ORCA-Drive controller results in a relatively low average speed. In contrast, the proposed Simplex-Drive architecture (with ORCA as BC and attention PPO as AC) can achieve higher success rates while maintaining desirable driving efficiency.
In addition, it can also be noticed that the proposed Simplex-Drive control architecture can deal with unexpected cases that do not occur in the DRL training process. Fig. \ref{orca_sim} illustrates an example scenario where crashed vehicles block the lane on which the ego-vehicle is driving. In this scenario, the proposed Simplex-Drive controller identifies potential unsafe states and switches to the ORCA-Drive safe control mode and thus avoids collision with crashed vehicles.

To further verify the safety of our switching logic, we conduct another experiment to simulate the circumstance that AC loses control. In such scenario, we use a ’dummy’ controller as AC which can only output constant acceleration actions (The ORCA-Drive controller is still used as BC). We evaluate the switches between the controllers and the ratio of the two control modes’ (AC and BC) active time among the overall operation time. 
During the experiment, the ratio of active time of BC increase from 23\% to 35\% when ’dummy’ controller performed slow acceleration action and from 51\% to 58\% when ’dummy’ controller performed aggressive action.
The results show that the ratio of BC controller’s active time increases as the the acceleration and traffic density grows, indicating that the switching logic can effectively predict potential unsafe states and take over the AC controller to ensure safety, while avoiding unnecessary frequent switching to maintain the control performance.

% \begin{figure}[h]
%     \centering
%         \includegraphics[width=0.4\textwidth]{figure/Ratio.PNG}
%         \vspace{-0.01 cm}
%     \caption{The controller switch ratio in circumstance that AC loses control. The IDLE means the AC output 0 acceleration and ACCELERATE means the AC output 0.5$m/s^2$ acceleration.}
%     \label{ratio}
% \end{figure}
 
\section{CONCLUSION}
In this paper, we have proposed Simplex-Drive, a reinforcement learning based control framework that can provide runtime safety assurance for autonomous vehicles. We have provided a set of formal proofs for the framework which can preserve safety properties all the time. Within the framework, the RL performance controller can help achieve superior control outputs; the ORCA safe controller can guarantee the drive safe conditions in challenging scenarios; the management unit can achieve smooth switching between two control modes. Simulation experiment results demonstrate that the proposed control system can provide safety guarantee without sacrificing control performance in terms of lane change success rate, lane change speed and trajectory smoothness in typical challenging autonomous driving scenarios. The proposed Simplex-Drive framework can be extended to various data-driven or model-based controllers, and serve as an attempt to develop a more general paradigm for building trustworthy AV systems. Our future work will focus on the implementation and validation of the proposed approach on physical vehicle platforms in real-world scenarios.


%%%%%%%%%%%%%%%%%%%%%%%%%%%%%%%%%%%%%%%%%%%%%%%%%%%%%%%%%%%%%%%%%%%%%%%%%%%%%%%%

%%%%%%%%%%%%%%%%%%%%%%%%%%%%%%%%%%%%%%%%%%%%%%%%%%%%%%%%%%%%%%%%%%%%%%%%%%%%%%%%

%%%%%%%%%%%%%%%%%%%%%%%%%%%%%%%%%%%%%%%%%%%%%%%%%%%%%%%%%%%%%%%%%%%%%%%%%%%%%%%%
%\section*{APPENDIX}


%\section*{ACKNOWLEDGMENT}

%The preferred spelling of the word ÒacknowledgmentÓ in America is without an ÒeÓ after the ÒgÓ. Avoid the stilted expression, ÒOne of us (R. B. G.) thanks . . .Ó  Instead, try ÒR. B. G. thanksÓ. Put sponsor acknowledgments in the unnumbered footnote on the first page.

%%%%%%%%%%%%%%%%%%%%%%%%%%%%%%%%%%%%%%%%%%%%%%%%%%%%%%%%%%%%%%%%%%%%%%%%%%%%%%%%

\bibliography{bibtex/reference.bib}{}
\bibliographystyle{IEEEtran}

\end{document}